\documentclass[10pt,twocolumn,letterpaper]{article}

\usepackage[pagenumbers]{cvpr} 

\definecolor{cvprblue}{rgb}{0.21,0.49,0.74}
\usepackage[pagebackref,breaklinks,colorlinks,allcolors=cvprblue]{hyperref}

\def\paperID{*****} 
\def\confName{CVPR}
\def\confYear{2026}

\title{Solution to the 10th ABAW Expression Recognition Challenge: A Robust Multimodal Framework with Safe Cross-Attention and Modality Dropout}

\author{Jun Yu$^{1}$,
 Naixiang Zheng$^{1,}$\footnotemark[1],
 Guoyuan Wang$^{1}$,
 Yunxiang Zhang$^{1}$,
 Lingsi Zhu$^{1}$\\
 Jiaen Liang$^{2}$,
 Wei Huang$^{2}$,
 Shengping Liu$^{2}$\\
$^{1}$University of Science and Technology of China\\
$^{2}$Unisound AI Technology Co., Ltd.\\
{\tt\small harryjun@ustc.edu.cn},\\ 
{\tt\small \{zhengnx, wgy2874849700, mesa, ls-zhu24\}@mail.ustc.edu.cn}\\ 
{\tt\small \{liangjiaen, huangwei, liushengping\}@unisound.com }
}

\begin{document}
\maketitle

\renewcommand{\thefootnote}{\fnsymbol{footnote}}
\footnotetext[1]{Corresponding author}

\begin{abstract}
    Emotion recognition in real-world environments is hindered by partial occlusions, missing modalities, and severe class imbalance. To address these issues, particularly for the Affective Behavior Analysis in-the-wild (ABAW) Expression challenge, we propose a multimodal framework that dynamically fuses visual and audio representations.  Our approach uses a dual-branch Transformer architecture featuring a safe cross-attention mechanism and a modality dropout strategy. This design allows the network to rely on audio-based predictions when visual cues are absent. To mitigate the long-tail distribution of the Aff-Wild2 dataset, we apply focal loss optimization, combined with a sliding-window soft voting strategy to capture dynamic emotional transitions and reduce frame-level classification jitter. Experiments demonstrate that our framework effectively handles missing modalities and complex spatiotemporal dependencies, achieving an accuracy of 60.79\% and an F1-score of 0.5029 on the Aff-Wild2 validation set.
\end{abstract}    
\section{Introduction}
\label{sec:intro}

Emotion recognition is central to human-computer interaction \cite{intro2,intro3,intro4}, psychological research and machine learning \cite{intro1}. Understanding human emotional states in real-world scenarios is necessary for developing empathetic systems \cite{intro5}. As a core component of affective computing, facial expression recognition enables machines to interpret human emotions, supporting applications in personalized education, mental health monitoring, customer behavior analysis, and social robotics \cite{intro4}.

While face detection and attribute recognition have seen considerable progress \cite{intro6,intro7}, accurately classifying human emotions remains difficult. Emotions are inherently subtle, ambiguous, and highly context-dependent, which complicates evaluation in practical applications. Furthermore, data collected in uncontrolled environments typically contains disturbances such as adverse lighting, partial occlusions, varying head poses, and cultural differences in expression, all of which limit model generalization. Traditional datasets, largely collected in controlled settings, fail to capture the full range of human emotional expression. Addressing these issues requires large-scale, diverse datasets that reflect natural interactions, as well as methods capable of integrating multimodal data.

The Affective Behavior Analysis in-the-wild (ABAW) workshop \cite{kollias2022abaw,kollias2023abaw,kollias2023abaw2,kollias20246th,kollias20247th} addresses these challenges by focusing on the multimodal analysis, modeling, and understanding of human emotions and behaviors in unconstrained environments \cite{kollias2021analysing,kollias2021affect,kollias2025emotions,kollias2025advancements,kollias2020analysing,kollias2019deep}. It hosts a competition with multiple tracks to help bridge the gap between controlled data and real-world applications. The 10th Competition on ABAW is split into  six Challenges :Valence-Arousal (VA) Estimation Challenge, Expression (EXPR) Recognition Challenge, Action Unit (AU) Recognition Challenge, Fine-Grained Violence Detection (VD) Challenge, Emotional Mimicry Intensity (EMI) Estimation Challenge, and Ambivalence/Hesitancy (AH) Video Recognition Challenge.

The Expression (EXPR) recognition challenge focuses on basic emotion classification using the Aff-Wild2 \cite{kollias2019expression} dataset. This dataset features a diverse range of subjects across different ages, genders, and ethnicities, recorded under varying lighting conditions, backgrounds, and head poses to reflect natural interaction scenarios. It contains 548 videos with approximately 2.7 million frames. Participants are required to perform frame-level expression recognition, classifying each frame into one of eight categories: six basic emotions (anger, disgust, fear, happiness, sadness, surprise), a neutral state, and an other category for emotional states outside these basics. Addressing the severe class imbalance and spatiotemporal feature extraction challenges in this track is critical for developing systems that can reliably interpret human emotions in the wild.

To address the aforementioned challenges of emotion recognition in real-world environments, we propose a novel multimodal model. By fusing visual and audio features, this model effectively improves the robustness of expression recognition in complex and unconstrained environments. The model incorporates a robust multimodal cross-attention mechanism based on a dual-branch Transformer \cite{transformer} architecture. This architecture independently extracts visual and audio contextual features and utilizes cross-attention blocks to facilitate inter-modality interactions. By introducing a learnable gating fusion mechanism \cite{gate}, the model adaptively balances the contributions of the unimodal context and the cross-modal fused features. To address the frequent occurrence of subjects exiting the field of view or experiencing severe occlusion in real-world scenarios, we develop a special attention module designed to handle missing modalities. This module introduces numerical protection and dynamic mask clearing logic into the calculation of cross-modal attention. When the system detects a complete absence of visual features within a specific window, it prevents the softmax function from generating invalid values. Relying on residual connections, the network gracefully degrades to rely entirely on the audio branch for decision-making in the absence of visual signals, substantially improving the fault tolerance of the system under extreme conditions.

We implement a sliding window approach alongside a soft voting inference strategy to capture the temporal dependencies inherent in extended video sequences during natural interactions. Overlapping sliding windows perform local spatiotemporal modeling. During the inference stage, a logit-based soft voting mechanism captures the dynamic transitions of emotional expressions and significantly reduces frame-level classification jitter by smoothing the predictions across adjacent windows. Furthermore, the framework integrates focal loss \cite{focal_loss} to mitigate the long-tail distribution bias characteristic of the Aff-Wild2 dataset. By applying focal loss as the optimization objective for the multimodal classifier, the framework dynamically reduces the weights of easily classified and high-frequency samples. This forces the model to concentrate on the challenging long-tail emotion samples, thereby enhancing both the generalization performance on minority categories and the overall evaluation metrics.
\section{Related Work}
\label{sec:related_work}

\subsection{In-the-Wild Affective Research}
Early facial expression recognition research relied on datasets from constrained laboratory environments \cite{past1,past2}. As the field shifted toward real-world applications, the need for 'in-the-wild' data became clear. Consequently, the Aff-Wild dataset \cite{zafeiriou2017aff} provided the first large-scale 'in-the-wild' benchmark for continuous valence and arousal estimation in unconstrained videos, initiating a series of related challenges. Because basic emotions do not cover the full range of human affective displays, the C-EXPR-DB dataset introduced extensive annotations for compound expressions, valence-arousal, and action units (AUs) \cite{kollias2023multi}. Expanding beyond facial expressions, the Diverse video Violence Database (DVD) provided large-scale, frame-level annotations to analyze anomalous behaviors like violence, improving upon previous coarse-grained datasets \cite{kollias2025dvd}.

Affective tasks (including classifying expressions, predicting AUs, and estimating valence-arousal) are closely related, making them well-suited for Multi-Task Learning (MTL). For example, FaceBehaviorNet jointly trains all three primary facial behavior tasks within a single network to improve emotion recognition \cite{kollias2019face}. However, MTL often struggles with partial or non-overlapping annotations across datasets, causing negative transfer. To address this, weakly-supervised coupling strategies, such as distribution matching and task-related label co-annotation, facilitate knowledge exchange between classification tasks without requiring fully overlapping labels \cite{kollias2024distribution}. These distribution matching techniques also couple compound expression recognition and AU detection in C-EXPR-NET to reduce negative transfer and improve performance \cite{kollias2023multi}.

The Behavior4All toolkit integrates face localization, expression recognition, AU detection, and valence-arousal estimation. It uses distribution matching to maintain performance and fairness across demographic groups \cite{kollias2024behaviour4all}. Building on these developments, our multimodal cross-attention framework dynamically fuses audio-visual cues to maintain robustness and handle missing modalities in unconstrained environments.

\subsection{EXPR in the ABAW Competition}

The Affective Behavior Analysis in-the-wild (ABAW) Expression (EXPR) recognition challenge has advanced multimodal emotion classification in unconstrained environments. Recent top-performing methods rely on pre-trained visual encoders to extract representations. For example, Zhang \etal. \cite{6thNeteast} pre-trained Masked Autoencoders (MAE) on facial datasets with up to 262 million images before fine-tuning on Aff-Wild2. Vision-Language models are also common; Lin \etal. \cite{6thxxx} paired a frozen CLIP image encoder with a trainable multilayer perceptron. They applied Conditional Value at Risk (CVaR) for class imbalance and Sharpness-Aware Minimization (SAM) to flatten the loss landscape. Similarly, Zhou \etal. \cite{8thCty} fine-tuned CLIP as a visual feature extractor for subsequent sequential learning modules.

Because facial expressions in videos are dynamic, temporal modeling and multi-modal fusion are standard in modern architectures. Zhou \etal. \cite{8thCty} captured temporal dependencies by integrating Temporal Convolutional Networks (TCN) with Transformer encoders. To move beyond static image features, Yu \etal. \cite{6thYu} used a dedicated temporal encoder to model relationships between neighboring frames. For modality fusion, Zhang \etal. \cite{6thNeteast} applied Transformer-based modules to integrate emotional cues from audio, visual, and textual streams. Yu \etal. \cite{8thYu} applied Global Channel-Spatial Attention (GCSA) to enhance unimodal features prior to proportional decision-level fusion.

Handling the data scarcity, severe class imbalance, and noisy annotations of the Aff-Wild2 dataset is another major focus. Yu \etal. \cite{6thYu} generated pseudo-labels from unlabeled face data using semi-supervised learning and applied debiased feedback learning to mitigate category imbalance. To optimize performance, Yu \etal. \cite{8thYu} designed coarse-fine granularity loss mechanisms that penalize severe emotion misjudgments. Savchenko \cite{8thHSE} filtered high-confidence frames using lightweight expression recognition models to avoid processing misleading data. Meanwhile, Zhang \etal. \cite{6thNeteast} applied scene-specific ensemble learning to improve overall accuracy.

While recent methods have improved performance, they still face limitations. Most multimodal fusion approaches assume that all modalities are continuously available. This assumption causes severe performance drops when subjects are occluded or leave the camera view, which frequently happens in unconstrained videos. In addition, the extreme long-tail distribution of naturalistic datasets like Aff-Wild2 biases models optimized with standard cross-entropy toward majority classes, reducing accuracy for minority classes. 
\section{Method}
\label{sec:method}
To address data noise, missing modalities, and severe class imbalance in unconstrained multimodal emotion recognition, we propose a robust end-to-end framework. Our approach consists of a two-stage feature extraction pipeline, a two-branch network with a safe cross-attention mechanism, and an inference strategy tailored for long video sequences.

\begin{figure*}[t]
  \centering
  \includegraphics[width=\linewidth]{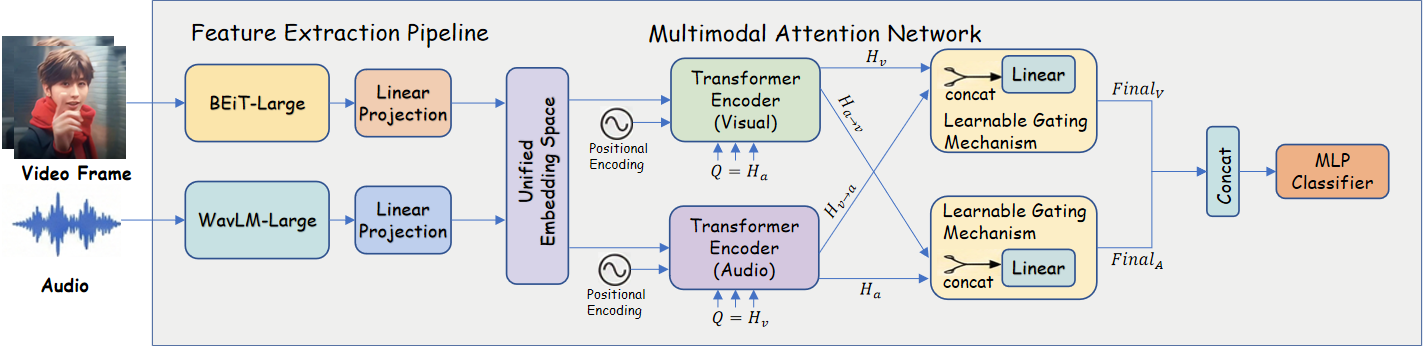}
  \caption{The proposed multimodal emotion recognition framework processes video and audio inputs through BEiT-Large and WavLM-Large, aligns them in a unified embedding space, and dynamically fuses the representations via a multimodal attention network for MLP classification.}
   \label{fig:onecol}
\end{figure*}

\subsection{Overview}
The proposed multimodal emotion recognition framework consists of five key components. First, a two-stage feature extraction and pre-training pipeline is employed to obtain robust visual and audio representations from unconstrained videos. Second, a multimodal attention network leverages cross-attention mechanisms and dynamic gating to align and fuse these unimodal features. Third, to address the issue of missing visual data, we introduce a modality dropout and safe attention mechanism that ensures stable performance even with incomplete modalities. Fourth, the network is optimized using a focal loss-based objective to mitigate the severe class imbalance inherent in the dataset. Finally, an inference strategy with overlapping sliding windows and post-processing is designed to produce smooth and continuous emotion predictions for long video sequences.

\subsection{Feature Extraction and Pre-training}
Learning high-dimensional representations directly from raw videos in unconstrained environments carries a high risk of overfitting. Therefore, we adopt a two-stage feature extraction pipeline combining pre-training and fine-tuning. For visual features, we use the BEiT-large \cite{beit} architecture as the backbone. To improve the model's generalization on basic facial expressions, we first construct a large-scale mixed static dataset from Raf-DB \cite{rafdb}, FERPlus \cite{ferplus}, and AffectNet \cite{affectnet}, keeping only high-quality manual annotations. This dataset covers the six basic expressions and the neutral state. After an initial fine-tuning phase on this mixed dataset, we apply domain-adaptive fine-tuning to the video frames of the target dataset (Aff-Wild2) to extract frame-level visual features, denoted as $V \in \mathbb{R}^{T \times d_v}$. For the audio modality, we separate the audio stream from the video and process it using WavLM-large, a model pre-trained on a large speech corpus. This allows us to capture acoustic prosody and subtle emotional fluctuations, yielding an audio feature sequence $A \in \mathbb{R}^{T \times d_a}$.

\subsection{Multimodal Attention Network}
To align and fuse multimodal features along the temporal dimension, our network uses unimodal encoders, cross-attention blocks, and a gating fusion module. First, linear projections map the visual and audio features into a shared hidden dimension $d_{model}$, adding sinusoidal positional encodings to preserve temporal order:

\begin{equation}
H_{v}^{(0)} = V W_v + PE, \quad H_{a}^{(0)} = A W_a + PE
\end{equation}

Next, Transformer encoders based on multi-head self-attention process these features to produce intra-modal contextual representations, $H_v$ and $H_a$. To enable cross-modal interaction, we apply cross-attention branches in both directions:

\begin{equation}
H_{v \rightarrow a} = \text{CrossAttn}(Q=H_v, K=H_a, V=H_a)
\end{equation}

\begin{equation}
H_{a \rightarrow v} = \text{CrossAttn}(Q=H_a, K=H_v, V=H_v)
\end{equation}

A learnable gating mechanism then dynamically regulates the information flow during fusion. Taking the visual branch as an example, a linear layer and a Sigmoid activation function compute the gating weight $G_v$ from the concatenated features:

\begin{equation}
G_v = \sigma(W_{gv} [H_v ; H_{v \rightarrow a}] + b_{gv})
\end{equation}

The final fused feature for the visual branch is $F_v = G_v \odot H_v + (1 - G_v) \odot H_{v \rightarrow a}$. The fused audio feature $F_a$ is obtained similarly. After concatenating $F_v$ and $F_a$, an MLP processes the joint representation to output the eight-class prediction.

\subsection{Modality Dropout and Safe Attention Mechanism}
In natural interactions, subjects often face occlusions or temporarily leave the camera view, leading to missing visual data. To improve robustness against such conditions, we introduce two mechanisms at the training and architectural levels. During training, a modality dropout strategy randomly masks the visual input within a batch with probability $p$, setting the visual window of selected samples to $v_{\text{mask}}=0$. This simulates real-world visual signal loss and prevents over-reliance on a single modality. Architecturally, if a complete absence of visual features is detected in a window, we temporarily unmask the first token of the sequence to allow the attention mechanism's forward pass to complete, and then manually set the attention output to zero ($\text{Attn\_output} = 0$). Combined with the residual connection ($Q = \text{LayerNorm}(Q + \text{Dropout}(\text{Attn\_output}))$), the layer automatically simplifies to $Q = \text{LayerNorm}(Q)$. This ensures the model fully retains the pure audio features during the $A \rightarrow V$ interaction, allowing it to maintain classification performance even without visual cues.

\subsection{Optimization Objective}
Standard cross-entropy loss is easily biased by majority classes. To mitigate the severe long-tail distribution in Aff-Wild2, we use focal loss \cite{focal_loss} for sequence-level optimization:

\begin{equation}
\mathcal{L}_{focal} = \frac{1}{\sum_{i=1}^{N} \mathbb{I}(y_i \neq -1)} \sum_{i=1}^{N} - (1 - p_{t, i})^\gamma \log(p_{t, i})
\end{equation}

where $p_{t, i}$ is the predicted probability of the true class, and the focusing parameter $\gamma$ is set to $2.0$. By down-weighting high-confidence samples, this loss function encourages the model to focus on difficult, minority-class samples. Additionally, we explicitly ignore invalid frames (labeled as $-1$) during loss computation to prevent gradient noise.

\subsection{Inference Strategy and Post-processing}
Because emotions transition continuously, our inference strategy for long videos relies on soft voting and temporal smoothing over overlapping sliding windows. We slice the input sequence using a window size of $W=64$ and a stride of $S=8$. For overlapping segments, rather than applying hard label voting, we average the predicted logits across all windows covering a given frame. Finally, we apply median filtering with a kernel size of $k=11$ to the frame-level predictions. This post-processing step reduces transient classification jitter caused by local noise or sudden modality shifts while preserving the boundaries of emotional states.
\section{Experiments}
\label{sec:experiments}

\subsection{Datasets and Pre-training Strategy}
\label{subsec:datasets}
We employ a multi-stage training strategy to extract visual features. First, we combine manually annotated data from Raf-DB, FERPlus, and AffectNet to construct a large-scale static dataset covering the six basic expressions and the neutral state. We use this mixed dataset to fine-tune several mainstream vision models before performing domain-adaptive fine-tuning on the target Aff-Wild2 dataset. To evaluate different visual backbones for real-world expression recognition, we compare ResNet50 \cite{resnet}, EfficientNetV2-M \cite{effnetv2}, MAE-ViT-Base \cite{maevit}, and BEiT-large on the Aff-Wild2 validation set, with results shown in Table \ref{tab:backbone_comp}.

\begin{table}[htbp]
    \centering
    \caption{Comparison of different visual backbones on the Aff-Wild2 validation set.}
    \label{tab:backbone_comp}
    \begin{tabular}{lcc}
        \toprule
        Visual Backbone & Accuracy & F1-Score \\
        \midrule
        ResNet50 & 0.4531 & 0.3345 \\
        EfficientNetV2-M & 0.4921 & 0.3932 \\
        MAE-ViT-Base & 0.4924 & 0.3686 \\
        BEiT-large & \textbf{0.5421} & \textbf{0.4268} \\
        \bottomrule
    \end{tabular}
\end{table}

BEiT-large achieves the best performance. Compared to convolutional networks like ResNet50 and EfficientNetV2-M, Transformer-based models are better suited for capturing global facial context. Furthermore, unlike MAE, which also uses masked image modeling, BEiT is pre-trained by predicting discrete visual tokens. This objective encourages the network to learn higher-level semantic representations rather than performing pixel-level reconstruction. Given its strong generalization, we adopt the fine-tuned BEiT-large as our default visual feature extractor.

\subsection{Multimodal Feature Extraction}
\label{subsec:feature_extraction}
For the visual modality, we pass video frames through the fine-tuned BEiT-large model and extract the pooled feature vectors from the final hidden layer. For the audio modality, we extract the audio stream from the original video using MoviePy, resample it to $16$ kHz, and process it using a pre-trained WavLM-large \cite{wavlm} model to obtain the final hidden layer features. Because the video and audio streams have different sampling rates, we apply linear interpolation to the WavLM output. This resamples the audio features to temporally align with the video frames. Given a video with $T$ frames, the aligned audio feature representation has a shape of $(T, d_a)$, where $d_a$ is the output dimension of WavLM.

\subsection{Baseline Design and Modality Weight Analysis}
\label{subsec:baseline}
To analyze the relative contributions of the visual and audio modalities, we design a frame-level two-stream baseline model using multilayer perceptrons (MLPs). The visual and audio features are projected through separate non-linear MLPs, followed by decision-level weighted fusion using a coefficient $\lambda$:

\begin{equation}
\text{Output} = \lambda V_{\text{out}} + (1 - \lambda) A_{\text{out}}
\end{equation}

\begin{table}[htbp]
    \centering
    \caption{Baseline performance with different modality fusion weights ($\lambda$).}
    \label{tab:baseline_fusion}
    \begin{tabular}{lcc}
        \toprule
        Weight & Accuracy & F1-Score \\
        \midrule
        0.0 & 0.4265 & 0.3150 \\
        0.5 & 0.5656 & 0.4427 \\
        0.7 & \textbf{0.5698} & \textbf{0.4436} \\
        1.0 & 0.5542 & 0.4368 \\
        \bottomrule
    \end{tabular}
\end{table}

Although the baseline performs frame-level prediction, audio signals are continuous. Therefore, to construct the audio input for a given frame, we extract the audio tokens for that frame along with a temporal context window of $18$ preceding and succeeding frames, applying average pooling to obtain the final representation. We evaluate this baseline on the Aff-Wild2 validation set using $\lambda \in \{0.0, 0.5, 0.7, 1.0\}$. 
As shown in Table \ref{tab:baseline_fusion}, the F1-scores follow the order $F_1(0.7) > F_1(0.5) > F_1(1.0) > F_1(0.0)$.

These results highlight the necessity of multimodal fusion. The significant gap between $F_1(1.0)$ and $F_1(0.0)$ confirms that facial visual features are the dominant modality for expression recognition. However, the combined setting ($F_1(0.7)$) outperforms the vision-only setting ($F_1(1.0)$), indicating that audio provides essential supplementary cues. In unconstrained environments, tone and pitch can compensate for missing visual information caused by head turns or occlusions. Still, the fusion must remain visually dominated; equal weighting ($\lambda = 0.5$) degrades performance by introducing noise that dilutes the primary visual signals.

\subsection{Sliding Window Strategy and Class-Balanced Setup}
\label{subsec:sliding_window}
To model temporal dependencies in long video sequences, we apply a sliding window sampling strategy to the extracted features. We set the window size to $W=64$ and the stride to $S=8$. To maintain training stability, we discard any window where the proportion of missing labels (invalid frames annotated as $-1$) exceeds $25\%$. Each training instance consists of the visual features $v_{\text{in}}$ of size $W \times d_v$, audio features $a_{\text{in}}$ of size $W \times d_a$, a visual mask identifier $v_{\text{mask}}$, and the corresponding frame-level labels. Aff-Wild2 suffers from severe class imbalance. To mitigate this, we dynamically scale the focal loss using class weights based on the effective number of samples \cite{effective_number_samples}. By estimating the actual coverage of each class in the feature space, we down-weight the gradient contributions from majority classes, which improves recognition accuracy for long-tail emotion categories.

\subsection{Ablation Studies on Model Architecture}
\label{subsec:ablation}
\begin{table}[htbp]
    \centering
    \caption{Ablation studies on modality dropout probability ($p$), dimension ($d$), and the number of self-attention layers ($l$).}
    \label{tab:model_ablation}
    \begin{tabular}{ccccc}
        \toprule
        $p$ & $d$ & $l$ & Accuracy & F1-Score \\
        \midrule
        0.0 & 256 & 2 & 0.5677 & 0.4628 \\
        0.0 & 512 & 2 & 0.5820 & 0.4739 \\
        0.0 & 256 & 3 & 0.5824 & 0.4764 \\
        0.0 & 512 & 3 & 0.5730 & 0.4626 \\
        \midrule
        0.10 & 256 & 3 & \textbf{0.6079} & \textbf{0.5029} \\
        0.10 & 256 & 4 & 0.5981 & 0.4814 \\
        \midrule
        0.15 & 256 & 3 & 0.5815 & 0.4819 \\
        \midrule
        0.20 & 256 & 3 & 0.5935 & 0.4734 \\
        \bottomrule
    \end{tabular}
\end{table}

We conduct ablation studies to determine the optimal multimodal architecture and modality dropout configuration. We systematically vary the hidden dimension $d$ of the Transformer encoder, the number of self-attention layers $l$, and the visual modality dropout probability $p$. The results on the Aff-Wild2 validation set are shown in Table \ref{tab:model_ablation}.

Comparing setups with ($p > 0$) and without ($p = 0.0$) modality dropout demonstrates that randomly dropping visual features improves robustness. For a network with $d=256$ and $l=3$, introducing a dropout probability of $p=0.10$ increases the F1-score from $0.4764$ to $0.5029$ and pushes the accuracy above $0.60$. Simulating visual signal loss during training forces the model to utilize the safe attention mechanism, preventing over-reliance on the visual modality and ensuring smooth degradation to audio-based predictions when visual cues are compromised. However, increasing $p$ to $0.15$ or $0.20$ degrades performance due to the excessive loss of primary visual information.

Regarding network capacity, increasing either the depth ($l=2$ to $l=3$) or the width ($d=256$ to $d=512$) under the $p=0.0$ setting yields only marginal gains. Increasing both simultaneously ($d=512, l=3$) causes the F1-score to drop to $0.4626$. Similarly, with $p=0.10$ and $d=256$, adding a fourth layer ($l=4$) harms performance. These trends suggest that highly parameterized models are prone to overfitting on noisy, limited-scale datasets like Aff-Wild2. Consequently, a medium-capacity network ($d=256$, $l=3$) provides the best trade-off between representation power and generalization.
\section{Conclusion}
\label{sec:conclusion}
We presented a multimodal emotion recognition framework to address missing modalities, noisy data, and class imbalance. Our approach uses domain-adapted BEiT-large and WavLM-large for feature extraction, fused via a dual-branch Transformer with safe cross-attention. By incorporating a modality dropout strategy, focal loss, and sliding-window soft voting, the model dynamically adapts to temporal emotional transitions and visual occlusions.

Our experiments on the Aff-Wild2 dataset provide practical insights into multimodal affective computing. While vision is the dominant modality, audio offers essential supplementary cues. Simulating visual loss through moderate modality dropout ($p=0.10$) improves fault tolerance. However, increasing network depth or width degrades performance, indicating that highly parameterized models are prone to overfitting on the noisy and long-tailed data typical of in-the-wild datasets. Future work will explore large-scale self-supervised learning on unlabelled naturalistic videos to reduce the need for manual annotations and mitigate overfitting. 
{
    \small
    \bibliographystyle{ieeenat_fullname}
    \bibliography{main}
}


\end{document}